\documentclass{article}
\usepackage[legalpaper, margin=1in]{geometry}

\usepackage{graphicx}
\usepackage{amsmath}
\usepackage{amsfonts}
\usepackage{amsthm}
\usepackage{appendix}

\usepackage{array}
\usepackage[section]{placeins}
\usepackage{afterpage}

\newtheorem{lemma}{Lemma}
\newtheorem{claim}{Claim}

\usepackage{hyperref}
\newcommand{\footremember}[2]{%
   \footnote{#2}
    \newcounter{#1}
    \setcounter{#1}{\value{footnote}}%
}
\newcommand{\footrecall}[1]{%
    \footnotemark[\value{#1}]%
} 

\newcommand{\bx}{\mathbf{x}}
\newcommand{\cR}{\mathcal{R}}
\newcommand{\bw}{\mathbf{w}}
\newcommand{\by}{\mathbf{y}}
\newcommand{\bz}{\mathbf{z}}
\newcommand{\ncomm}[1]{}

\newcommand{\norm}[1]{\left\lVert#1\right\rVert}

\begin{document}

\title{Adiabatic Quantum  Feature Selection for Sparse Linear Regression }

\author{Surya Sai Teja Desu\footremember{first}{Indian Institute of Technology Hyderabad, India; \href{mailto:cs17b21m000002@iith.ac.in}{cs17b21m000002@iith.ac.in},\href{mailto:srijith@cse.iith.ac.in,mvp@cse.iith.ac.in}{\{srijith,mvp\}@cse.iith.ac.in}}, P.K. Srijith\footrecall{first}, M.V. Panduranga Rao\footrecall{first}, Naveen Sivadasan\footremember{second}{TCS Research, Hyderabad, India; \href{mailto:naveen.sivadasan@tcs.com}{naveen.sivadasan@tcs.com}}}
\date{}
\maketitle              % typeset the header of the contribution
\begin{abstract}
Linear regression is a popular machine learning approach to learn and predict real valued outputs or dependent variables from independent variables or features. In many real world problems, its beneficial to perform sparse linear regression to identify important features helpful in predicting the dependent variable. It not only helps in getting interpretable results but also avoids overfitting when the number of features is large, and the amount of data is small. The most natural way to achieve this is by using `best subset selection' which penalizes non-zero model parameters by adding $\ell_0$ norm over parameters to the least squares loss. However, this makes the objective function non-convex and intractable even for a small number of features.  This paper aims to address the intractability of sparse linear regression with $\ell_0$ norm using adiabatic quantum computing, a quantum computing paradigm that is particularly useful for solving optimization problems faster.  We formulate the $\ell_0$ optimization problem as a Quadratic Unconstrained Binary Optimization (QUBO) problem and solve it using the D-Wave adiabatic quantum computer. We study and compare the quality of QUBO solution on synthetic and real world datasets. The results demonstrate the effectiveness of the proposed adiabatic quantum computing approach in finding the optimal solution. The QUBO solution matches the optimal solution for a wide range of sparsity penalty values across  the datasets. 
% \keywords{adiabatic quantum computing  \and sparse linear regression \and feature selection}
\end{abstract}
\section{Introduction}

Most of the real world application of machine learning arising from various domains such as web, business, economics, astronomy and science involve solving regression problems~\cite{Montgomery2001IntroductionTL,Leatherbarrow1990UsingLA,Yatchew1998NonparametricRT,Wu2002ANA}. Linear regression is a popular machine learning technique to solve the regression problems where a real valued scalar response variable (dependent variable or output) is predicted using explanatory variables (independent variables or input features)~\cite{hastie01,bishop06}. It assumes the output variable to be an affine function of the input variables and learns the model parameters (weight coefficient parameters) from the data by minimizing the least squares error. In many practical applications, we are not only interested in learning such functions but also in understanding the importance of the input features in determining the output. Identifying relevant features helps in interpretability, which is important for many real-world applications. 
Moreover, these problems are often associated with a large number of input variables and consequently large number weight parameters. When the data is limited, least squares approach to learn the parameters will result in over-fitting and poor predictive performance~\cite{bishop06}. 

Sparse linear regression has been proposed to address the above limitations that are associated with least squares regression. It encourages sparse weight vectors by adding to the loss function, appropriate norms over the weight vectors. Using $\ell_0$ norm over the weight vector, which is known as the {\it{best subset selection method}}~\cite{hocking67,hastie01}, is an effective approach to identify features useful for linear regression. $\ell_0$ norm counts the number of non-zero components in the weight vector and favours solutions with smaller number of non-zero elements. Consequently, one can identify the important features as the ones with non-zero weight vectors. Best subset selection has been theoretically shown to achieve optimal risk inflation~\cite{foster94}. However,  solving an optimization problem involving $\ell_0$ norm is non-convex and intractable as the number of possible choices of non-zero elements in weight vectors is exponentially large. In fact, this is an NP-hard problem and can become intractable even for small values of data dimension and subset size~\cite{natarajan95}. Approaches such as greedy algorithm  based {\it Forward- and Backward-Stepwise Selection }~\cite{hastie01,zhang09} and heuristics-based integer  programming~\cite{konno09,miyashiro15}  were proposed to select subset of features for linear regression. However, they  are sub-optimal and  are not close to the optimal solution of  best subset selection. Instead of modelling sparsity exactly through $\ell_0$ norm, a common practice is to use its convex approximation  using $\ell_1$ norm, known as Lasso regression~\cite{tibshirani96}.  However, it will  lead to sub-optimal selection of features, and incorrect models as the shrinkage property can result in a  weight vector with many elements zero. It can also lead to a biased model as it heavily penalizes the weight coefficients, even the ones corresponding to the relevant and active features~\cite{mazumder11,friedman12}.  

In this paper, we propose to use adiabatic quantum computing technique for training sparse linear regression with $\ell_0$ norm. Given the NP-hardness of the problem, a tractable solution is not expected even with quantum computers. However, quantum computers have been found to be effective in speeding up training of machine learning algorithms. In particular, adiabatic quantum computing is found to be excellent in solving hard optimization problems~\cite{dwave}.  Adiabatic quantum computers like D-Wave 2000Q (from D-Wave)  can efficiently solve quadratic unconstrained binary optimization (QUBO) problems.  Machine learning algorithms relying on optimization techniques for parameter estimation can benefit from (adiabatic) quantum computing. It has been  effectively used to train machine learning models like deep belief networks (DBN)~\cite{Date2019ACH} and least squares regression~\cite{QAnnealingLLS,QlinearR,date2020quboML,wang2017quantum,elmahalawy2020classical}. While much of the previous work focuses only on least squares regression, in this paper we solve the $\ell_0$ regularized least squares regression problem.  We derive a Quadratic Unconstrained Binary Optimization (QUBO) problem for the same, and solve it using the D-Wave adiabatic quantum computer. We report an extensive evaluation of the performance of the QUBO approach on  synthetic  and  real-world  data. Our experimental results indicate that the QUBO based approach for solving large scale sparse linear regression problems shows promise. Our main contributions are summarized as follows.

\begin{enumerate}
    \item We propose an adiabatic quantum computing approach for the best subset selection based sparse linear regression problem. 
    \item We formulate the sparse linear regression problem as a Quadratic Unconstrained Binary Optimization (QUBO) problem. Our approach uses QUBO to solve best subset selection optimally and computes the corresponding regression coefficients using the standard least squares regression. 
    \item We conduct experiments on the DWave quantum computer using synthetic and real-world datasets to demonstrate the performance of the proposed approach. For a wide range of sparsity penalty values, the QUBO solution matches the optimal solution across the datasets.
\end{enumerate}

\section{Background}
In this section, we give a brief overview of the adiabatic quantum computing 
approach for solving optimization problems. We refer the interested reader to
the excellent survey by Venegas-Andraca et al~\cite{dwave}.

\subsection{Optimization Problems}
Given a set $S$ of $n$ elements, denote by $\mathcal{P}(S)$ its power set.
Consider a function $f:\mathcal{P}(S)\rightarrow \mathbb{R}$. The \emph{combinatorial
optimization} problem (we consider, without loss of generality, the
minimization problem) is to find $P\in\mathcal{P}(S)$ such that
$f(P)=\min_{P_i\in \mathcal{P}(S)}(f(P_i))$. In the absence of any structure, these optimization problems are NP-hard.

\subsection{Three Formulations of Optimization}

It is possible to formulate these (NP-hard) minimization problems as
\begin{itemize} 
\item  Minimization of pseudo-Boolean functions:

Consider a function $f:\{0,1\}^n\rightarrow \mathbb{R}$ such that 
\[
f({\bf x}) = \sum_{I\subseteq [n]} \gamma_I \prod_{x_j\in I}x_j,
\]
where $\gamma_I\in \mathbb{R}$. The degree of this multilinear polynomial is $max\{I\}$. The problem is to find 
the ${\bf x}$ for which $f({\bf x})$ is minimum.

\item Quadratic Unconstrained pseudo-Boolean Optimization (QUBO):

Consider a function
\[
f_B({\bf x}) = \sum_{i=1}^n\sum_{j=1}^i \beta_{ij} x_ix_j,
\]
where $x_i\in\{0,1\}$ and $\beta_{ij}\in \mathbb{R}$. Notice that this is a degree two polynomial. Again, the problem in this case is to find ${\bf x}$ for which $f({\bf x})$ is minimum.

\item The Ising Model:

Given {\bf(i)} a graph $G=(V,E)$, real valued weights $h_{v}$ assigned to each vertex $v\in V$, and real valued weights $J_{uv}$ assigned to each
edge $(u,v)\in E$,  {\bf(ii)} A set of Boolean variables called \emph{spins}: $S=\{s_1,\ldots, s_n\}$ where $s_{u}\in \{-1,1\}$, the spin $s_v$
corresponding to the vertex $v$ and {\bf(iii)} an \emph{energy function}
\[
E(S)= \sum_{v\in V} h_vs_v + \sum_{(u,v)\in E}J_{uv}s_us_v.
\]
The problem here is to find an assignment to the spins that minimizes the energy function.
\end{itemize}

It is well known that any pseudo-Boolean function minimization problem can be reformulated as a QUBO minimization in polynomial time \cite{Ishikawa,dwave}.
Further, a QUBO problem can be very easily converted to an optimization problem in the Ising Model in a straightforward manner, namely by
using $s_i = 2x_i - 1$.

\subsection{Adiabatic Quantum Computation}
We now briefly discuss the adiabatic quantum computing paradigm. The
interested reader is referred to \cite{farhi2000quantum}. In what follows, we assume a basic
knowledge of quantum mechanics, in particular, the notion of a Hamiltonian
operator.

To solve an optimization problem, the system is initially prepared in
the ground state of an initial Hamiltonian $H_I$. Naturally, this ground
state should be ``easy to prepare". The ground state of
a ``final" Hamiltonian $H_F$ encodes the solution to the optimization
problem. Let the system, prepared in the ground state of $H_I$, evolve
as per the following \emph{interpolating} Hamiltonian for a time duration
$T$.
\[
H\left(\frac{t}{T}\right)= \left(1-\frac{t}{T}\right)H_I + \frac{t}{T}H_F.
\]

The adiabatic gap theorem says that if the $T\geq O(\frac{1}{g_{min}^2})$,
where $g_{min}$ is the spectral gap, the system will stay in the
ground state of the Hamiltonian with high probability. Therefore, with a
high probability, the system will end up in the ground state of the final
Hamiltonian, the solution of the optimization that we are searching for. This evolution is analogous to classical simulated annealing; the difference being that due to phenomena like superposition and entanglement that are peculiar to quantum mechanics, the system \emph{tunnels} through local peaks instead of going over them. In what follows, we will use the \emph{quantum annealing} interchangeably with adiabatic quantum computing.

\subsection{Ising Model Implementation}
Quantum computers like the Quantum Processing Unit(QPU) of D-Wave, that are based on the adiabatic quantum computing, typically 
implement an Ising model~\cite{TransverseIsing}. Let $G_{QC}$ be the graph  with qubits as vertices and connectors as edges. 
For a vertex $i$ in such a graph, one can talk of $h_i$ the magnetic field on qubit $i$ and $J_{ij}$ the so-called coupling strength
between qubits $i$ and $j$. Further, let $\sigma^z_i$ be the Pauli z matrix acting on qubit $i$. Then,

The problem Hamiltonian $H_F$ has the Ising formulation:
\begin{equation}\label{QIsing}
H_F = \sum_{i} h_i\sigma^z_i + \sum_{i<j} J_{ij} \sigma^z_i \sigma^z_j. 
\end{equation}

While the graph structure in the Ising optimization formulation can be generic, it may not be isomorphic 
to the $G_{QC}$, which for D-wave QPU's current implementation is a chimera graph. However, it is possible to obtain a minor embedding
in a subgraph of $G_{QC}$ corresponding to $G$ \cite{dwave}. Then, it is possible to obtain $h_i$  and $J_{ij}$ from the Ising formulation of the 
minimization problem. The ground state of the Hamiltonian $H_F$ then corresponds the spin assignment to $S$ that minimizes 
the function in Eq (\ref{QIsing}). 

This yields the following approach for using  quantum annealing to solve a minimization problem: 
\begin{enumerate}
    \item Pose the minimization problem as a pseudo-Boolean
function minimization problem
    \item Convert the pseudo-Boolean function to a QUBO formulation
    \item The QUBO formulation is then converted to an
Ising model formulation on the native QPU
\end{enumerate}
The minimization is achieved through adiabatic evolution.

\subsection{Linear Regression}

We consider a regression problem with input-output pair $(\mathbf{x},y)$, where $\mathbf{x} \in \mathcal{R}^d$ and $y \in \mathcal{R}$ (for e.g. house price prediction, grade prediction). We assume the training dataset to be $\mathcal{D} = \{\mathbf{x}_i, y_i \}_{i=1}^N$. The goal is to learn a function $f:\cR^d \rightarrow \cR$ with good generalization performance from the training dataset.   Linear regression is one of the widely used approach for regression problems which assumes y is a linear function of  $\mathbf{x}$. Non-linear regression can also be modelled through linear regression framework by considering higher order powers of $\mathbf{x}$. In either case, the functional form is linear in terms of the  parameters $\mathbf{w} \in \mathcal{R}^d$ such that  $y = \mathbf{w}^{T}\mathbf{x}$.

In practice,  one cannot find a function which passes through all the data points. Hence, the  function is learnt to be as close as possible to the observations. Training in linear regression involves learning the parameters $\bw$ such that squared error between actual values and the values given by the function $f$ is minimum. The resulting optimization problem would be to find $\mathbf{w}$ such that it minimizes the following squared error loss \cite{bishop06}: 
\begin{equation}
    \min_{\mathbf{w}}\sum_{i=1}^N(y_i-\mathbf{w}^{T}\mathbf{x}_i)^2.
\end{equation}
The solution to the above problem can be computed in closed form as 
\begin{equation}\label{eq:lsr}
    \mathbf{w} = (X^{T}X)^{-1}X^{T} \by,
\end{equation}
where we assume $X$ is a $N \times d$ matrix formed by stacking \ncomm{} the input data as $N$ row vectors, $\by$ is an $N$ dimensional column vector of the training data outputs. 
We assume that the matrix $X$ is column normalized where each column of $X$ is divided by its $\ell_2$ norm.

However, on datasets with large dimensions and limited data, learning the parameters by minimizing the squared loss alone can result in learning a  complex model which will overfit on the data. Under these circumstances, one could achieve a very low training error but a high test error and consequently poor generalization (predictive) performance.  For instance, in predicting tumour based on genetic information, the input dimension is in the order of thousands while the number of samples is often in the order of hundreds. In such problems,  only a few dimensions are relevant and contribute to the output prediction. Instead of solving the regression problem using all the input features, solving using the best subset of features could lead to functions with better generalization performance. Moreover, this can help in identifying features which are relevant to the regression problem and this aids in interpret ability.   This is ideally achieved by adding a $\ell_0$ regularization term over the parameters to the least squares loss. $\ell_0$ norm regularizer
is defined as
$$\norm{\mathbf{\bw}}_0 = \sum_{j=1}^d \mathrm{I} \{\bw_j \neq 0 \}$$.
This counts the number of non-zero elements in the weight vector and consequently it determines the dimensions which are active and relevant. The resulting regularized loss function to estimate $\mathbf{w}$ is given as
\begin{equation}
    \min_{\mathbf{w}}\sum_{i=1}^N(y_i-\mathbf{w}^{T}\mathbf{x}_i)^2+ \lambda \|\mathbf{w}\|_{0}.
    \label{eqn:regls}
\end{equation}
However,  $\ell_0$ norm is discontinuous, and the above optimization problem is known to be NP-Hard.  However, it gives the best possible solution to sparse linear regression~\cite{hastie01}.

\section{Proposed Methodology}

Adiabatic quantum computing system in D-Wave requires the problem to be specified as a QUBO expression \cite{dwave}. The QUBO expression is then internally converted to an Ising model and is fed to the D-Wave quantum computing system.  We propose an approach to express the regularized least squares regression problem with $\ell_0$ norm  given by (\ref{eqn:regls}) as a QUBO problem. 
Existing approaches (eg~\cite{date2020quboML}) for the related problem of standard least square regression assumes a bounded word length $k$ for the elements of $\mathbf{w}$ in order to construct a QUBO formulation and the number of logical qubits needed depends on $k$ as well as the dimension $d$. We attempt to construct a QUBO formulation to solve the subset selection optimally. We then solve the values of the selected elements of $\mathbf{w}$ classically. In this way, we avoid any assumption on the word length for the elements of $\mathbf{w}$. Furthermore, the number of logical qubits used in our case is only a function of the dimension $d$.
We introduce a binary vector (selection vector) $\mathbf{z} \in \{0,1\}^d$, with  $z_j$ determining whether dimension $j$ is selected. The best subset selection problem can be solved as an optimization problem involving both the  selection vector $\mathbf{z}$ and the weight vector $\bw$.
We first rewrite the original formulation in (\ref{eqn:regls}) using the selection vector as follows
\begin{equation}\label{eq:r}
\min_{\mathbf{w,z}}\sum_{i=1}^{N}(y_{i}-\mathbf{w}^{T}{\bx'}_i)^2+\lambda \|\textbf{z}\|_0,
\end{equation}
where vector $\mathbf{x'}_i = \mathbf{x}_i\circ\mathbf{z}$, is the element-wise product between the input vector $\mathbf{x_i}$ and the selection vector $\mathbf{z}$. 
\ncomm{The optimization problem can be solved as a co-ordinate descent optimization problem where the optimization is performed with respect the selection vector $\mathbf{z}$ and the weight vector $\bw$ alternatively.} For any arbitrarily fixed $\bz$, solution for $\mathbf{w}$ can be obtained by first solving   
\begin{equation}\label{eq:w}
    \mathbf{w'} = (X'^{T}X')^{-1}X'^{T}\mathbf{y},
\end{equation}
where $X'$ is the $N \times d'$ submatrix of the $N \times d$ matrix $X$, where $X'$ is obtained by retaining only the $\norm{z}_0 = d'$ columns of $X$ corresponding to non-zeros in the vector $\mathbf{z}$. Vector $\mathbf{w'}$ is simply a projection of the $d'$ components of the target vector $\mathbf{w}$ corresponding to the $d'$ non-zeros in $\mathbf{z}$. Since, the remaining components of $\mathbf{w}$ are all zeros, $\mathbf{w}$ is trivially obtained from $\mathbf{w'}$.

We use Lemma \ref{lem:approx} (Section \ref{sec:approx}), to obtain the following first order approximation $(X'^T X')^{-1} \approx \alpha(2I - \alpha X'^T X')$, where $\alpha = 2/(d+1)$. Substituting this in (\ref{eq:w}) yields  $\mathbf{w'} = \alpha(2I-\alpha X'^{T}X')X'^{T}\mathbf{y}$. We can also view $X'$ as the $N \times d$ matrix $X \times  \mathbf{diag}(\mathbf{z})$, where the $\mathbf{diag}(\mathbf{z})$ is a diagonal matrix whose diagonal corresponds to the vector $\mathbf{z}$. Clearly, in this $X'$, columns corresponding to the zeros in $\mathbf{z}$ are all zeros. With this definition of $X'$, it is straightforward to verify that, we can directly write the expression for $\mathbf{w}$ in place of $\mathbf{w'}$ as
\begin{equation} \label{eq:final}
    \mathbf{w} ~=~  \alpha(2I-\alpha X'^{T}X')X'^{T}\mathbf{y}.
\end{equation}

\subsection{QUBO Formulation}
We derive an equivalent  QUBO formulation for (\ref{eqn:regls}) using (\ref{eq:r}) and the expression for $\bw$ from (\ref{eq:final}). 
Using (\ref{eq:final}), we obtain $i$th component $w_i$ of $\mathbf{w}$ as

\[
w_i=\alpha \left(\left(2-\alpha z_i^2\sum_{p=1}^N x_{pi}^2 \right)\sum_{k=1}^{N}x_{ki}z_iy_{k}- \sum_{j=1, j \neq i }^{d} \sum_{p=1}^Nx_{pi}z_i.x_{pj}z_j\sum_{k=1}^{N}x_{kj}z_jy_{k}\right).
\]

Recalling that for each column $i$ of $X$,  $\sum_{p=1}^N x_{pi}^2$ = 1, we expand the above expression to obtain
\begin{eqnarray}
w_i &=& \alpha(2-\alpha z_i^2)\sum_{k=1}^{N}x_{ki}z_iy_{k} ~~-~~ \alpha^2\sum_{j=1, j \neq i }^{d} \sum_{p=1}^Nx_{pi}z_i.x_{pj}z_j\sum_{k=1}^{N}x_{kj}z_jy_{k}\nonumber\\
&=& z_{i}\left(\alpha(2-\alpha)\sum_{k=1}^{N}x_{ki}y_{k}  ~~-~~ \alpha^2\sum_{j=1, j \neq i }^{d}z_{j} \sum_{p=1}^N x_{pi} x_{pj}\sum_{k=1}^{N}x_{kj}y_{k}\right). 
\end{eqnarray}
The last step follows because $z_{i}^c = z_{i}$ for binary $z_{i}$,  for all positive integer $c$. Substituting the above expression for $w_{i}$, we can rewrite (\ref{eq:r}) as

\begin{eqnarray}
  & &  \min_{\mathbf{z}}~\sum_{t=1}^N \left(y_t - \sum_{i=1}^dw_ix_{ti}z_i\right)^2+\lambda\sum_{r=1}^dz_r \nonumber\\
&=&    \min_{\mathbf{z}}~\sum_{t=1}^N\left(y_t -
    \sum_{i=1}^d\left(\alpha(2-\alpha)\sum_{k=1}^{N}x_{ki}y_{k}- \alpha^2\sum_{j=1, j \neq i }^{d}z_j \sum_{p=1}^Nx_{pi} x_{pj}\sum_{k=1}^{N}x_{kj}y_{k}\right)x_{ti}z_i\right)^2+\lambda\sum_{r=1}^dz_r. \nonumber
\end{eqnarray}

Denoting the elements of $X^TX$ by $p_{ij}$ and the elements of $X^Ty$ by
$q_i$, the above minimization can be reformulated as

\begin{eqnarray}
& &   \min_{\mathbf{z}}~\sum_{t=1}^N\left(y_t -
   \sum_{i=1}^d\left(\alpha(2-\alpha)q_{i}-\alpha^2\sum_{j=1, j \neq i }^{d}z_j p_{ij}q_j\right)x_{ti}z_i\right)^2+\lambda\sum_{r=1}^dz_r \nonumber\\
&=&    \min_{\mathbf{z}}~\sum_{t=1}^N \left(y_t -
    \sum_{i=1}^d\left(\alpha(2-\alpha)q_{i}x_{ti}z_{i}-\sum_{j=1, j \neq i }^{d}\alpha^2p_{ij}q_jx_{ti}z_{j}z_{i}\right)\right)^2+\lambda\sum_{r=1}^dz_r. \nonumber
\end{eqnarray}

Letting $b(t)_{ij} = \alpha^2(p_{ij}q_{j}x_{ti}+p_{ji}q_{i}x_{tj})$ and  $q'_{i} = \alpha(2-\alpha)q_{i}$, the above minimization becomes
\begin{equation}\label{ref_eqn}
\min_{\mathbf{z}}~\sum\limits_{t=1}^{N}\left(y_t-\sum\limits_{i=1}^{d}q'_{i}x_{ti}z_{i} ~+\sum\limits_{1\leq i<j\leq d}b(t)_{ij}z_{i}z_{j}\right)^2 ~+~ \lambda \sum\limits_{r=1}^{d}z_{r}.
\end{equation}

Expanding (\ref{ref_eqn}) further and again simplifying using $z_{i}^c = z_{i}$, we finally obtain
\begin{equation*}\label{qubo}
\min_{\mathbf{z}}\left(y+\sum\limits_{i=1}^{d}\left(e_{i}+\lambda\right)z_{i}+\sum\limits_{1\leq i<j\leq d}f_{ij}z_{i}z_{j}+\sum\limits_{1\leq i<j<k \leq d}g_{ijk}z_{i}z_{j}z_{k} + \sum\limits_{1\leq i<j<k<l \leq d}h_{ijkl}z_{i}z_{j}z_{k}z_{l}\right),
\end{equation*}
where 
\begin{eqnarray}
e_{i} &=& \sum\limits_{t=1}^{N}\Big({q'_{i}}^{2} x_{ti}^{2} ~-~2q'_{i}x_{ti}y_t\Big),\nonumber\\
f_{ij} &=& \sum\limits_{t=1}^{N}\Big(b(t)_{ij}^2+2\Big(q'_iq'_jx_{ti}x_{tj}-\big(q'_ix_{ti}+q'_jx_{tj}-y_t\big)b(t)_{ij}\Big)\Big),\nonumber\\ 
g_{ijk} &=& \sum\limits_{t=1}^{N}\Big(2\Big(b(t)_{ij}b(t)_{ik}+b(t)_{ij}b(t)_{jk}+b(t)_{ik}b(t)_{jk} \nonumber \\ 
&& ~~~~- q'_{k}x_{tk}b(t)_{ij} - q'_{i}x_{ti}b(t)_{jk} - q'_{j}x_{tj}b(t)_{ik}\Big)\Big),\nonumber\\
h_{ijkl} &=& \sum\limits_{t=1}^{N}2 \Big(b(t)_{ij}b(t)_{kl} ~+~b(t)_{ik}b(t)_{jl} ~+~ b(t)_{il}b(t)_{jk}\Big), \textrm{~~~and~} y  =  \sum\limits_{t=1}^{N}y_{t}^2.\nonumber
\end{eqnarray}

We make use of the in-built method `make\_quadratic()' in Ocean SDK provided by D-Wave~\cite{ocean} to obtain a quadratic equivalent (QUBO expression) of the above binary optimization problem. D-wave uses efficient approaches \cite{dwave} to turn higher degree terms to quadratic terms by introducing auxiliary binary variables. In our case, it follows that the final QUBO formulation would involve $O(d^4)$ logical qubits.

The QPU finds an optimal solution for $\bz$. An optimal solution for $\bz$ corresponds to an optimal feature selection. Once an optimal choice of features is known, the problem of solving $\mathbf{w}$ reduces to the standard least squares regression problem (Eq (\ref{eq:lsr})), which can be solved efficiently using any of the existing methods.

\subsection{Approximation of \texorpdfstring{$(X^TX)^{-1}$}{(XTX){-1}}}
\label{sec:approx}
We derive a bound on $(X^TX)^{-1}$, which is a direct adaptation of the well-known Neumann series \cite{neumann}. However, we include the whole proof here for completeness. We recall that every column of the $N \times d$ matrix $X$ is $\ell_2$ normalized. We assume that $X^TX$ is full rank. Clearly, following Lemma also holds true for any $(X'^TX')^{-1}$, where $X'$ is a submatrix of $X$ obtained by retaining some $d' \le d$ columns of $X$.

\begin{lemma}
\label{lem:approx}
$
 (X^TX)^{-1}  ~=~ \alpha  \lim_{n \to \infty} \sum_{i=0}^n (I- \alpha X^TX)^i
$
for positive $\alpha \le \frac{2}{d+1}$.
\end{lemma}
\begin{proof}
From Lemma \ref{lem:approx}, a $k$th order approximation of $(X^TX)^{-1}$ is given by \newline
$
 (X^TX)^{-1}  ~\approx~ \alpha  \sum_{i=0}^k (I - \alpha X^TX)^i
$, where $\alpha = 2/(d+1)$. In order to prove the Lemma, we first show the following Claim.

\end{proof}

\begin{claim}
\label{cl:eigen}
Eigenvalues of $X^T X$ are in the range $[0, d]$.
\end{claim}

\begin{proof}
Since $X^TX$ is a $d \times d$ Gram matrix, $X^TX$ is a symmetric positive semi-definite matrix. Hence, all eigenvalues of $X^T X$ are non-negative.
Let $U [\lambda] U^T$ be the eigendecomposition of $X^TX$ where $[\lambda]$ is the diagonal matrix of eigenvalues of $X^TX$.
We note that $U$ is an orthonormal $d \times d$ matrix where $UU^T = I$. Thus, for any vector $\mathbf{a}$, 
$$
\norm{\mathbf{a} U^T}^2_2 = \norm{\mathbf{a}}^2_2.
$$

Let $U = [\mathbf{u}_1, \ldots, \mathbf{u}_d]$, where $\mathbf{u}_i$s are column vectors. Similarly, let $U^T = [\mathbf{v}_1, \ldots, \mathbf{v}_d]$. Let $X = A U^T$, where $A = X U$. Let the $N \times d$ matrix $A$ be viewed as stacking of $N$ row vectors $\mathbf{a}_1, \ldots, \mathbf{a}_N$ in the same order.

Since, columns of $X$ are normalized and since $X = A U^T$, it is easy to verify that for any $j \in \{1, \ldots, d\}$,
$
\sum_{i=1}^N \langle \mathbf{a}_i, \mathbf{v}_j \rangle^2 = 1
$.
It follows that, 
$$
\sum_{j=1}^d\sum_{i=1}^N \langle \mathbf{a}_i, \mathbf{v}_j \rangle^2 = d.
$$

Interchanging the summations, we obtain
$
 \sum_{i=1}^N \sum_{j=1}^d\langle \mathbf{a}_i, \mathbf{v}_j \rangle^2 = d
$.

The inner summation is $\norm{\mathbf{a}_i U^T}^2_2$. Recalling that $\norm{\mathbf{a}_i U^T}^2_2 = \norm{\mathbf{a}_i}^2_2$, it follows that
$
 \sum_{i=1}^N \norm{\mathbf{a}_i}^2_2  = d
$. In other words, $\sum_{i=1}^N \sum_{j=1}^d {a}_{ij}^2  = d$,
where $a_{ij}$ is the entry at $i$th row and $j$th column of matrix $A$, which is also the $j$th component of the vector $\mathbf{a}_i$.

Recalling that $X =  AU^T$ and that $X^TX = U [\lambda] U^T$, we have
$$
X^TX = UA^TAU^T = U[\lambda]U^T.
$$ 

In other words, 
$
A^T A = [\lambda]
$. The $j$th diagonal entry of $A^TA$ is given by $\sum_{i=1}^N a_{ij}^2$. However, recalling that 
 $\sum_{i=1}^N \sum_{j=1}^d {a}_{ij}^2  = d$
, it follows that $j$th diagonal entry of $A^TA$ satisfies 
$$
\sum_{i=1}^N a_{ij}^2 ~\le~ \sum_{i=1}^N \sum_{j=1}^d {a}_{ij}^2  ~=~ d.
$$

It follows that every entry of $[\lambda]$ is in the range $[0, d]$.
\end{proof}

\begin{proof}[Lemma \ref{lem:approx}]
Let the eigendecomposition of $X^TX = U [\lambda] U^T$. It follows that the eigendecomposition of 
$$
\alpha X^TX = U [\alpha \lambda] U^T.
$$ where $[\alpha \lambda]$ is the scalar multiplication of $\alpha$ with the eigenvalue diagonal matrix $[\lambda]$.

Let 
$
Y ~=~ \alpha X^TX
$. From Claim \ref{cl:eigen}, it follows that all eigenvalues of a full rank $X^TX$ are in $(0, d]$. Consequently,  all eigenvalues of $Y$ are in $(0, 2)$ for $\alpha \le 2/(d+1)$. 

Let
$
B ~=~ I - Y
$. Since eigenvalues of $Y$ are in $(0, 2)$, it follows  that the eigenvalues of $B$ are in $(-1, 1)$. Since all eigenvalues of $B$ are strictly less than $1$ in absolute value, it follows that $\lim_{n \to \infty} B^n = 0$.
In other words, $\lim_{n \to \infty} \sum_{i=0}^n B^i$ is a convergent series. Following Neumann series gives

$$
\lim_{n \to \infty} (I - B) \sum_{i= 0}^n B^i   ~= ~
 \lim_{n \to \infty} \left( \sum_{i= 0}^n B^i - \sum_{i=0}^{n} B^{i+1} \right) = 
\lim_{n \to \infty} (I - B^{n+1})  = I .
$$

Thus,
$
Y^{-1} ~=~ (I - B)^{-1} ~=~  \lim_{n \to \infty} \sum_{i=0}^n B^i
$
Recalling $Y = \alpha X^TX$, where $\alpha$ is a positive scalar, and $B = I-Y$, we obtain
$$
\alpha^{-1} (X^TX)^{-1}  ~=~  \lim_{n \to \infty} \sum_{i=0}^n (I - Y)^i.
$$

The result follows by observing that,
$
 (X^TX)^{-1}  ~=~ \alpha  \lim_{n \to \infty} \sum_{i=0}^n (I-Y)^i.
$
\end{proof}

\section{Experimental Results}
For experimental evaluation of our approach, we consider both synthetic datasets as well as real world data. The QUBO formulation is run on D-Wave 2000Q quantum computer which has 2048 qubits.  For a given dataset, we run the corresponding QUBO instance on D-Wave for multiple runs. Each run outputs a feature selection $\mathbf{z}$. The corresponding $\mathbf{w}$ and the objective function value is obtained from Eq (\ref{eq:w}) and Eq (\ref{eqn:regls}) respectively. The final solution  is chosen as the one that minimizes the objective function. We compare the quality of QUBO solution with optimal solution for this NP-hard problem computed by exhaustive search in the classical setting. This exhaustive search is performed over all possible feature selections to find the solution that minimizes Eq (\ref{eqn:regls}). Here again, for a given feature selection, $\mathbf{w}$ is obtained from Eq (\ref{eq:w}).

Synthetic datasets were generated using randomly chosen $X$ and a randomly chosen sparse vector $\mathbf{w}$. The outputs are obtained as $\mathbf{y} = X^\top \mathbf{w}$.  We generate a separate synthetic input data for input dimensions in the range $5, \ldots, 10$. For each of these dimensions, the number of samples $N$ in the input data is fixed as $3000$. For each of the resulting six input datasets, we use QUBO to solve Eq~(\ref{eqn:regls}) for 5 different $\lambda$ values. Table 1 summarises QUBO results. For each input data and $\lambda$ combination, the table gives the $\ell_0$ norm of $\mathbf{w}$, which corresponds to the cardinality of the selected features, computed using QUBO and the optimal $\mathbf{w}$ computed classically by exhaustive search. Columns 6 and 7 of the table give the values of the objective function  Eq~(\ref{eqn:regls}) corresponding to $\mathbf{w}$ obtained through the classical solution and QUBO respectively.  
A set of $\lambda \times d$ values were used so as to reduce the sparsity penalty $\lambda$ for increasing $d$. This balances the contributions of the regression gap and the sparsity penalty in the objective function.   The last five rows of Table 1 show results for 500 runs, while the rest are for 100 runs.
 We can observe  that the cardinality of $\mathbf{w}$, i.e. the number of features selected, is the same for the QUBO solution and classical solution for most values of $\lambda$ across all the synthetic datasets. Similarly, the objective function values are also very close for the QUBO and  classical solution. For data with input dimension 10, we also experimented with 500 QUBO runs. These are included as the last 5 rows in the table. As shown in the table, increasing the number of runs further improved the performance of the QUBO solver in finding  optimal solutions. For higher values of $\lambda$, we observe an increased gap between the QUBO solution and the optimal solution. This is partly due to the fact that higher values of $\lambda$
induces sparser $\mathbf{z}$. In such situations, choosing $\alpha$ closer to $2/(d'+1)$ in Eq~(\ref{eq:final}), where $d'$ is the cardinality of the optimal feature set, can yield a better inverse approximation compared to the existing $2/(d+1)$ value. However, the value of $d'$ is not known \emph{a priori}. Hence, using a higher order inverse approximation or running QUBO with different permissible values of $\alpha$ are potential ways to mitigate this issue.

\begin{table}[hbt!]
       
%\end{table}

%\begin{table}
    \centering
    \begin{tabular}{|m{1.1cm}|m{1.1cm}|m{1cm}|m{1.1cm}|m{1cm}|m{1.67cm}|m{1.67cm}|m{1.5cm}|m{1.5cm}|}
    \hline
        Number of points (N) & Number of features (d) & $\lambda \times $d & $\norm{\mathbf{w}}_0$ Classical & $\norm{\mathbf{w}}_0$ QUBO & Objective value Classical & Objective value  QUBO& Preproce- ssing time (sec) & Processing time (sec) \\
        \hline
3000	&	5	&	10	&	1	&	2	&	2.43	&	4	&	0.3802	&	1.7997	\\
3000	&	5	&	1	&	2	&	2	&	0.4	&	0.4	&	0.3015	&	1.463	\\
3000	&	5	&	0.1	&	2	&	2	&	0.04	&	0.04	&	0.2921	&	1.8042	\\
3000	&	5	&	0.01	&	2	&	2	&	0.004	&	0.004	&	0.3087	&	1.5339	\\
3000	&	5	&	0.001	&	2	&	2	&	0.0004	&	0.0004	&	0.3158	&	1.7676	\\
\hline
3000	&	6	&	10	&	1	&	3	&	3.1993	&	5	&	0.5311	&	1.6666	\\
3000	&	6	&	1	&	2	&	3	&	0.4542	&	0.5	&	0.5421	&	1.6226	\\
3000	&	6	&	0.1	&	3	&	3	&	0.05	&	0.05	&	0.5286	&	1.6272	\\
3000	&	6	&	0.01	&	3	&	3	&	0.005	&	0.005	&	0.5807	&	1.5697	\\
3000	&	6	&	0.001	&	3	&	3	&	0.0005	&	0.0005	&	0.5341	&	1.6663	\\
\hline
3000	&	7	&	10	&	2	&	3	&	2.9767	&	4.2858	&	0.8584	&	1.9038	\\
3000	&	7	&	1	&	2	&	3	&	0.4053	&	0.4286	&	0.985	&	1.8351	\\
3000	&	7	&	0.1	&	3	&	3	&	0.0429	&	0.0429	&	0.9155	&	1.8839	\\
3000	&	7	&	0.01	&	3	&	3	&	0.0043	&	0.0043	&	0.8463	&	1.8471	\\
3000	&	7	&	0.001	&	3	&	3	&	0.0005	&	0.0005	&	0.8983	&	1.8971	\\
\hline

%3000	&	8	&	10	&	2	&	4	&	2.658	&	5	&	1.3872	&	2.7443	\\
%3000	&	8	&	1	&	3	&	4	&	0.3886	&	0.5128	&	1.3208	&	2.8005	\\
%3000	&	8	&	0.1	&	4	&	4	&	0.05	&	0.0631	&	1.3319	&	2.7639	\\
%3000	&	8	&	0.01	&	4	&	4	&	0.005	&	0.0181	&	1.3219	&	2.9072	\\
%3000	&	8	&	0.001	&	4	&	4	&	0.0005	&	0.0005	&	1.3149	&	2.92	\\
%\hline
3000	&	9	&	10	&	2	&	4	&	2.3813	&	4.4573	&	2.0121	&	3.7735	\\
3000	&	9	&	1	&	3	&	4	&	0.3468	&	0.4445	&	1.9516	&	4.4245	\\
3000	&	9	&	0.1	&	4	&	4	&	0.0445	&	0.0445	&	2.0444	&	4.4954	\\
3000	&	9	&	0.01	&	4	&	4	&	0.0045	&	0.0045	&	1.9706	&	4.5724	\\
3000	&	9	&	0.001	&	4	&	4	&	0.0005	&	0.0005	&	1.9625	&	4.5866	\\
\hline
3000	&	10	&	10	&	3	&	5	&	3.1369	&	5.0124	&	3.8412	&	7.2634	\\
3000	&	10	&	1	&	4	&	5	&	0.4128	&	0.5	&	3.8206	&	7.5119	\\
3000	&	10	&	0.1	&	5	&	6	&	0.05	&	0.1573	&	3.6595	&	7.2924	\\
3000	&	10	&	0.01	&	5	&	5	&	0.005	&	0.0174	&	3.6433	&	7.3122	\\
3000	&	10	&	0.001	&	5	&	6	&	0.0005	&	0.0127	&	3.6286	&	7.3667	\\
\hline
\hline
3000	&	10	&	10	&	3	&	5	&	3.1369	&	5	&	3.0189	&	6.6345	\\
3000	&	10	&	1	&	4	&	5	&	0.4128	&	0.5	&	3.1579	&	7.244	\\
3000	&	10	&	0.1	&	5	&	5	&	0.05	&	0.05	&	3.0992	&	7.0419	\\
3000	&	10	&	0.01	&	5	&	5	&	0.005	&	0.005	&	2.8799	&	7.3774	\\
3000	&	10	&	0.001	&	5	&	5	&	0.0005	&	0.0005	&	2.8831	&	7.1387	\\
        \hline
    \end{tabular}
    %\centering
     \caption{Experimental results on synthetic datasets for different $d$ values. 
   % A set of $\lambda \times d$ values were used so as to reduce the sparsity penalty $\lambda$ for increasing $d$. This balances the contributions of the regression gap and the sparsity penalty in the objective function.  The last five rows show results for 500 runs, while the rest are for 100 runs.
   }
    \label{tab:my_label}
    
    \centering
    \begin{tabular}{|m{1.1cm}|m{1.1cm}|m{1cm}|m{1.1cm}|m{1cm}|m{1.67cm}|m{1.67cm}|m{1.5cm}|m{1.5cm}|}
    \hline
        Numb-er of points (N) & Num- ber of features (d) & $\lambda$ & $\norm{\mathbf{w}}_0$ Classical & $\norm{\mathbf{w}}_0$ QUBO & Objective value Classical & Objective value  QUBO & Preproc- essing time (sec) & Processing time (sec) \\
        \hline
        442	&	10	&	10000	&	6	&	6	&	11561403.16	&	11615190.26	&	0.52024	&	7.71478	\\
442	&	10	&	1000	&	8	&	8	&	11502623.87	&	11519510.11	&	0.52523	&	8.0403	\\
442	&	10	&	100	&	9	&	7	&	11494877.38	&	11554655.73	&	0.51418	&	7.75863	\\
442	&	10	&	10	&	10	&	7	&	11493995.03	&	11511792.48	&	0.52244	&	7.51821	\\
442	&	10	&	1	&	10	&	8	&	11493905.03	&	11511518.11	&	0.51872	&	7.47994	\\
\hline
    \end{tabular}
    %\centering
    \caption{Results on the real-world Diabetes dataset for 100 QUBO runs.}
    \label{tab:diab}
\end{table}

\afterpage{\FloatBarrier}

We also measured the mean squared error (MSE) for the optimal classical and the QUBO settings on the train and test data.  The test error was measured using a held-out test data of 1000 points generated separately for each $d$, using the same distribution as the training data.  In particular, we measured the absolute gap between the test errors of classical and QUBO as well as between the train errors.  For values of $d  \times \lambda \le 1$, the gap was less than $10^{-6}$.  For $d  \times \lambda = 10$ case, the gap was of the order of $10^{-3}$.

The preprocessing time is the time taken to create the QUBO formulation using the D-wave libraries. The processing time includes the network time to 
upload the problem instance to D-wave, network time to download the solution and the execution time on the D-Wave quantum computer. 
We used the default annealing time 
(20 micro seconds) provided in the D-Wave Sampler API (SAPI) servers \cite{ocean} and with this, the execution time for each run 
was around 11 milli seconds. The reported processing time is almost entirely the data upload and download overhead. Processing time is averaged over the 
number of QUBO runs.

Table 2 shows the QUBO performance for the real-world Diabetes dataset~\cite{diabetes}. The number of QUBO runs here is 100. For this dataset, smaller $\lambda$ values did not induce any sparsity even for the classical optimal solution, due to high values of regression error term in the objective function. Hence, we chose larger $\lambda$ values to induce sparsity. Here again, we observe that the objective function values are similar for the QUBO solution and the classical solution, and increasing the number of QUBO iterations can possibly narrow the gap further.

\section{Future Directions}
We believe that this work is an important step towards applying Adiabatic Quantum Computing for efficiently solving practical and large scale sparse linear regression problems. 
    This paper leaves open several future directions to explore. 
    Our experiments were limited by a restricted access to the D-wave infrastructure. Experiments involving larger problem instances needs to be done to measure the quantum advantage. Moreover, experimentation with different annealing schedules and durations are likely to yield faster convergence to the optimal solution.
    Different ways of reducing the effect of inverse approximation can be explored. Examples are higher order approximation and approximations using multiple feasible values of $\alpha$. A practical direction would be to apply the QUBO based regression model on real world data and compare with other models to measure generalization error.

\bibliographystyle{splncs04}

\begin{thebibliography}{10}
\providecommand{\url}[1]{\texttt{#1}}
\providecommand{\urlprefix}{URL }
\providecommand{\doi}[1]{https://doi.org/#1}

\bibitem{bishop06}
Bishop, C.M.: Pattern recognition and machine learning. Springer (2006)

\bibitem{QAnnealingLLS}
Borle, A., Lomonaco, J.: Analyzing the quantum annealing approach for solving
  linear least squares problems. In: WALCOM (2019)

\bibitem{ocean}
D-Wave: D-wave's ocean software, \url{https://docs.ocean.dwavesys.com/}

\bibitem{QlinearR}
Date, P., Potok, T.: Adiabatic quantum linear regression. arXiv:2008.02355
  (2020)

\bibitem{Date2019ACH}
Date, P., Schuman, C.D., Patton, R.M., Potok, T.: A classical-quantum hybrid
  approach for unsupervised probabilistic machine learning. In: Future of
  Information and Communication Conference. pp. 98--117 (2019)

\bibitem{date2020quboML}
Date, P., Arthur, D., Pusey-Nazzaro, L.: Qubo formulations for training machine
  learning models. arXiv:2008.02369  (2020)

\bibitem{diabetes}
Diabetes dataset : https://www4.stat.ncsu.edu/~boos/var.select/diabetes.html

\bibitem{elmahalawy2020classical}
El-Mahalawy, A.M., El-Safty, K.H.: Classical and quantum regression analysis
  for the optoelectronic performance of ntcda/p-si uv photodiode.
  arXiv:2004.01257  (2020)

\bibitem{farhi2000quantum}
Farhi, E., Goldstone, J., Gutmann, S., Sipser, M.: Quantum computation by
  adiabatic evolution (2000)

\bibitem{foster94}
Foster, D.P., George, E.I.: The risk inflation criterion for multiple
  regression. The Annals of Statistics p. 1947--1975 (1994)

\bibitem{friedman12}
Friedman, J.H.: Fast sparse regression and classification. International
  Journal of Forecasting, 28(3) p. 722--738 (2012)

\bibitem{hastie01}
Hastie, T., Tibshirani, R., Friedman, J.: The Elements of Statistical Learning.
  Springer New York Inc. (2001)

\bibitem{hocking67}
Hocking, R.R., Leslie, R.N.: Selection of the best subset in regression
  analysis. Technometrics 9(4) p. 531--540 (1967)

\bibitem{Ishikawa}
Ishikawa, H.: Transformation of general binary mrf minimization to the
  first-order case. IEEE Trans. Pattern Anal. Mach. Intell.  \textbf{33}(6),
  1234--1249 (2011)

\bibitem{TransverseIsing}
Kadowaki, T., Nishimori, H.: Quantum annealing in the transverse ising model.
  Phys. Rev. E  \textbf{58},  5355--5363 (Nov 1998)

\bibitem{konno09}
Konno, H., Yamamoto, R.: Choosing the best set of variables in regression
  analysis using integer programming. Journal of Global Optimization pp.
  273--282 (2009)

\bibitem{Leatherbarrow1990UsingLA}
Leatherbarrow, R.: Using linear and non-linear regression to fit biochemical
  data. Trends in biochemical sciences  \textbf{15} (1990)

\bibitem{mazumder11}
Mazumder, R., Friedman, J.H., Hastie, T.: Sparsenet: Coordinate descent with
  non-convex penalties. J. American Statistical Association, 106 (495) (2011)

\bibitem{miyashiro15}
Miyashiro, R., Takano, Y.: Subset selection by {M}allows' {C}p: A mixed integer
  programming approach. Expert Systems with Applications, p. 325--331. (2015)

\bibitem{Montgomery2001IntroductionTL}
Montgomery, D., Peck, E.A., Vining, G.G.: Introduction to Linear Regression
  Analysis. Wiley (2001)

\bibitem{natarajan95}
Natarajan, B.K.: Sparse approximate solutions to linear systems. SIAM journal
  on computing, p. 227--234 (1995)

\bibitem{neumann}
Riesz, F., Sz.~Nagy, B.: Functional Analysis. F. Ungar Pub. Co. (1955)

\bibitem{tibshirani96}
Tibshirani, R.: Regression shrinkage and selection via the lasso. Journal of
  the Royal Statistical Society. Series B (Methodological) p. 267--288 (1996)

\bibitem{dwave}
Venegas-Andraca, S.E., Cruz-Santos, W., McGeoch, C., Lanzagorta, M.: A
  cross-disciplinary introduction to quantum annealing-based algorithms.
  Contemporary Physics  (2018)

\bibitem{wang2017quantum}
Wang, G.: Quantum algorithm for linear regression. Physical review A
  \textbf{96}(1),  012335 (2017)

\bibitem{Wu2002ANA}
Wu, B., Tseng, N.: A new approach to fuzzy regression models with application
  to business cycle analysis. Fuzzy Sets Syst.  \textbf{130},  33--42 (2002)

\bibitem{Yatchew1998NonparametricRT}
Yatchew, A.: Nonparametric regression techniques in economics. Journal of
  Economic Literature  \textbf{36},  669--721 (1998)

\bibitem{zhang09}
Zhang, T.: Adaptive forward-backward greedy algorithm for sparse learning with
  linear models. Adv. in Neural Information Processing Systems p. 1921--1928
  (2009)

\end{thebibliography}

\end{document}